\documentclass{article}

\usepackage[preprint]{neurips_2025}

\usepackage{enumitem}

\usepackage{url}

\usepackage[dvipsnames, svgnames, x11names,table,xcdraw]{xcolor}
\usepackage{graphicx}
\usepackage{booktabs}
\usepackage[linesnumbered,ruled,vlined]{algorithm2e}
\usepackage{multirow} 
\usepackage{subfig}

\usepackage{hyperref}
\hypersetup{
    colorlinks=true, 
    linkcolor=red, 
    citecolor=blue,
    urlcolor=blue  
}

\definecolor{HeaderGray}{HTML}{D0CECE}
\definecolor{Stripe}{HTML}{E7E6E6}
\definecolor{OursBg}{HTML}{FFF2CC}

\usepackage[accsupp]{axessibility}

\author{
    Siquan Huang$^1$,
    Yijiang Li$^2$,
    Ningzhi Gao$^1$,
    Xingfu Yan$^{3*}$,
    Leyu Shi$^1$,\\
    \textbf{Ying Gao$^{1*}$} \\
    $^1$School of Computer Science and Engineering, South China University of Technology \\
    $^2$Department of Electrical and Computer Engineering, University of California San Diego \\
    $^3$School of Computer Science, South China Normal University
}

\title{BackdoorIDS: Zero-shot Backdoor Detection for Pretrained Vision Encoder}

\begin{document}

\maketitle

\begin{abstract}
Self-supervised and multimodal vision encoders learn strong visual representations that are widely adopted in downstream vision tasks and large vision–language models (LVLMs). 
% However, training these encoders on internet-scale data is computationally expensive and exposes them to data poisoning, including maliciously injected backdoors. This risk is further amplified in practice, where downstream users often rely on third-party pretrained encoders with uncertain provenance, increasing vulnerability to backdoor attacks.
However, downstream users often rely on third-party pretrained encoders with uncertain provenance, exposing them to backdoor attacks.
% In this work, we propose a simple yet effective zero-shot, inference-time backdoor detection method for these vision encoders operating, termed \textbf{BackdoorIDS}. 
% Existing defenses either depend on access to the training dataset or require suitable auxiliary data, which are impractical for addressing a compromised vision encoder. 
In this work, we propose \textbf{BackdoorIDS}, a simple yet effective zero-shot, inference-time backdoor samples detection method for pretrained vision encoders. \textbf{BackdoorIDS} is motivated by two observations: \textit{Attention Hijacking} and \textit{Restoration}. Under progressive input masking, a backdoor image initially concentrates attention on malicious trigger features. Once the masking ratio exceeds the trigger’s robustness threshold, the trigger is deactivated, and attention rapidly shifts to benign content. This transition induces a pronounced change in the image embedding, whereas embeddings of clean images evolve more smoothly across masking progress.
\textbf{BackdoorIDS} operationalizes this signal by extracting an embedding sequence along the masking trajectory and applying density-based clustering such as DBSCAN. An input is flagged as backdoor if its embedding sequence forms more than one cluster.
% \textbf{BackdoorIDS} builds on two observations,  \textbf{Attention Hijacking} and \textbf{Restoration}. By progressively masking the input, backdoor images  will first have attention  largely concentrated on malicious features. As the masking ratio grows and exceeds the trigger’s robustness threshold, the trigger deactives and attention quickly shifts to benign features. This creates a large change in the image embedding, while benign ones remain uniformly across the masking process.
% first compute similarities score of the first $k$ image representations in the masking sequence. the average local density of the top-k embeddings to determine the clustering radius, then scale it and apply density-based clustering to all embeddings. 
% Extensive experiments demonstrate that \textbf{BackdoorIDS} significantly outperforms existing methods across a wide range of attack types, datasets, and models. Notably, it serves as a plug-and-play solution compatible with various encoder architectures, including CNN, ViT, CLIP, and LLaVA-1.5 in a purely zero-shot manner during inference time computation.
Extensive experiments show that \textbf{BackdoorIDS} consistently outperforms existing defenses across diverse attack types, datasets, and model families. Notably, it is a plug-and-play approach that requires no retraining and operates fully zero-shot at inference time, making it compatible with a wide range of encoder architectures, including CNNs, ViTs, CLIP, and LLaVA-1.5.
  % \keywords{Vision Encoder \and Backdoor Defense \and Zero-shot \and Inference-time}
\end{abstract}

\section{Introduction}

Self-supervised learning (SSL) and multimodal pretraining on massive internet-scale unlabeled data have become the dominant paradigm for learning general-purpose foundation models \cite{gui2024survey, jaiswal2020survey}. By leveraging large-scale, weakly curated visual and multimodal data, visual representation learners acquire rich and transferable visual representations that serve as foundational components for a wide range of downstream vision applications \cite{chen2020simple, li2022more,chen2020big}. These strong learners also play a central role in modern large vision-language models (LVLMs) \cite{liu2024improved, zhang2025unified, zhang2024pixels, zhu2023minigpt}, where they serve as the vision encoder, providing the visual perception necessary for multimodal reasoning and generation.

 % In particular, vision encoders trained through SSL effectively understand visual-semantic relationships and generate high-quality visual representations \cite{chen2020simple, chen2020big}. Thus, these encoders not only serve as effective feature extractors for downstream classifier construction, but also contribute to the development of large vision-language models (LVLMs) \cite{liu2024improved, zhu2023minigpt}.

% However, the performance of SSL vision encoders is highly dependent on large amounts of unlabeled training data. Besides, training large-scale encoders also requires substantial computational resources and is prohibitively expensive, as model parameters grow \cite{brown2020language,kaddour2023challenges}. As a result, regular users face significant barriers due to limited data and hardware, and must rely on externally provided pretrained encoders, such as CLIP released by OpenAI \cite{radford2021learning}. This reliance on third parties introduces considerable security risks, particularly from backdoor attacks \cite{chen2017targeted,gu2017badnets}. 

However, training vision encoders on internet-scale data enlarges the attack surface for data poisoning and backdoor injection \cite{carlini2024poisoning}. At the same time, the computational resources required to train these models become increasingly prohibitive as their scale grows \cite{brown2020language, kaddour2023challenges}, pushing most users to rely on third-party pretrained encoders with uncertain provenance. Unfortunately, this dependence on unreliable encoders exposes users to the risk of backdoor attacks \cite{chen2017targeted,Huang_2023_ICCV,gu2017badnets}. Adversaries can either release compromised encoders directly or poison web-scale training data \cite{carlini2024poisoning}, yielding backdoor models that appear benign on clean inputs yet can be reliably steered by trigger-bearing samples, thereby subverting downstream systems \cite{saha2022backdoor,jia2022badencoder, 10852410, tao2024distribution, li2023embarrassingly, liang2024badclip, liu2025stealthy, zhang2024data, chen2025backdooring, yuan2023you}. Moreover, backdoor attacks have evolved from using conspicuous patch triggers to employing adaptive, optimization-based designs, making triggered inputs increasingly difficult to distinguish \cite{li2023embarrassingly,yuan2023you,liang2024badclip}. They also alter the training process, further complicating the identification even in the embedding space \cite{tao2024distribution,liang2024badclip, ma-etal-2025-jailbreaking, liu2025stealthy, chen2025backdooring}.

To address these sophisticated attacks, a range of defense strategies has been proposed. Backdoor sample detection methods \cite{feng2023detecting,pan2023asset,rong2025backdoor,doan2023defending,hou2025dede} typically identify anomalies in the distribution of image embeddings, which can effectively separate backdoor inputs from clean ones. However, these methods \cite{feng2023detecting,pan2023asset,rong2025backdoor} heavily rely on access to the original pretraining data, which is often impractical for third-party pretrained vision encoders. To address this limitation, more recent paradigms perform inference-time detection with a small auxiliary dataset \cite{doan2023defending,hou2025dede}. Nevertheless, their effectiveness is highly sensitive to distributional alignment between the auxiliary data and the encoder’s pretraining data distribution. Without access to the original pretraining corpus, their performance is also limited in practical.
This raises a critical question: \textit{Can backdoor inputs be reliably detected at inference time in a fully zero-shot setting?}

In this setting, we assume access only to the test input and a potentially compromised vision encoder, with no auxiliary data. We first examine how backdoor affect vision encoders by mechanistically analyzing their impacts on attention and image embeddings.
Prior work \cite{rong2025backdoor} shows that patch triggers often localize attention (concentrate its attention on a small region, as shown in (b) of Fig. \ref{attention_map}), but this cue does not generalize to blended triggers that are distributed across the whole image (see (c) of Fig. \ref{attention_map}). Instead, we posit a shared property of both patch-based and blended attacks: attention is minimally distributed across benign features and predominantly concentrated on trigger-related features. We refer to this phenomenon as \textbf{Attention Hijacking}.
Under hijacking, small perturbations leave the embedding nearly unchanged as long as the trigger remains active. When perturbations exceed the trigger’s robustness threshold, the trigger deactivates; attention rapidly reorients toward benign features, producing an abrupt embedding shift, which we call \textbf{Attention Restoration}. Consequently, under progressive masking, clean inputs exhibit smooth, gradual changes in attention and embeddings, whereas backdoor inputs undergo a sharp transition once masking suppresses the trigger.

Motivated by this phenomenon induced by backdoors, we propose \textbf{BackdoorIDS}, a zero-shot inference-time detector for pretrained vision encoders. We progressively mask the input to generate a sequence of images and encode them to obtain an embedding trajectory. Clean inputs yield smoothly varying embeddings with roughly uniform adjacent distances, whereas backdoor inputs produce an initial tight cluster (\textbf{Attention Hijacking}) followed by an abrupt jump when the trigger deactivates (\textbf{Attention Restoration}). We set the DBSCAN radius using the scaled average local density of the top-$k$ embeddings, and cluster all embeddings with DBSCAN \cite{ester1996density}: clean inputs form one cluster, while backdoor inputs split into multiple clusters. We classify by cluster count (one = clean; $>1$ = backdoor). Across datasets, attacks, and encoder architectures (including LVLMs), \textbf{BackdoorIDS} outperforms baselines with strong robustness to noise and JPEG compression \cite{wallace1991jpeg}.

Comprehensively, we summarize our main contribution as follows:

\begin{itemize}[nosep,leftmargin=*]
% \begin{itemize}
    \item We uncover \textbf{Attention Hijacking} and \textbf{Attention Restoration}, showing that backdoor and clean inputs exhibit markedly different embedding dynamics under the same progressive masking process. These distinctions enable inference-time backdoor samples detection without prior knowledge or auxiliary data.
    \item Based on these observations, we propose \textbf{BackdoorIDS}, a zero-shot inference-time backdoor samples detector for pretrained vision encoders that is plug-and-play and requires neither additional data nor training.
    \item Extensive experiments across datasets, attacks, and models show that BackdoorIDS consistently outperforms prior methods and is robust across various vision encoders (e.g., CNN \cite{he2016deep}, ViT \cite{dosovitskiy2020image}, CLIP \cite{radford2021learning}, LLaVA-1.5 \cite{liu2024improved}) and tasks (e.g., classification, captioning).
\end{itemize}

\section{Related Work}
\subsection{SSL Vision Encoder}
There are three mainstream types of SSL approaches: contrastive learning (CL), multimodal contrastive learning (MCL), and masked autoencoders (MAE) \cite{gui2024survey}. CL \cite{chen2020simple, chen2020big, chuang2020debiased} learns representations by encouraging the embeddings of the same input to be close, while pushing apart those of different inputs. In MCL \cite{mustafa2022multimodal, radford2021learning, yuan2021multimodal}, images and text are mapped into a shared embedding space, where matching image-text pairs are pulled together, and mismatched pairs are pushed apart. CLIP \cite{radford2021learning} is a representative form of MCL using natural language supervision trained on a vast dataset consisting of 400 million image-text pairs sourced from the internet. 
MAE \cite{he2022masked, tong2022videomae, woo2023convnext} learns by reconstructing missing parts of input data to get image features. After pretraining with these methods, vision encoders can be used for downstream tasks such as image classification \cite{deng2009imagenet, netzer2011reading, stallkamp2012man} and captioning \cite{chen2015microsoft}.

\subsection{Backdoor Attacks in Vision Encoder}
The pioneering work on backdoor attacks in SSL vision encoders includes SSLBackdoor \cite{saha2022backdoor} and BadEncoder \cite{jia2022badencoder}. BadEncoder \cite{jia2022badencoder} introduces a novel paradigm for injecting backdoors into pretrained vision encoders, enabling their activation in downstream classification tasks. Within the contrastive learning framework, it minimizes the distance between the embeddings of backdoor and target samples to inject the backdoor. Building on this, CorruptEncoder \cite{zhang2024data} enhances backdoor effectiveness by disrupting the correlation between the trigger and background features, causing the model to focus on the trigger. NA \cite{chen2025backdooring} constructs a backdoor sample dataset through noise alignment, improving the robustness of backdoor attacks. To achieve the stealthiness of the backdoor, CTRL \cite{li2023embarrassingly} adds perturbations in the high-frequency band of the sample to serve as the trigger, ensuring the trigger remains invisible. Drupe \cite{tao2024distribution} narrows the gap between the embedding distributions of backdoor and clean samples, making it difficult to distinguish between them and further complicating backdoor detection. Besides, BadViT \cite{yuan2023you} focuses on Vision Transformers (ViT) \cite{dosovitskiy2020image}, optimizing a specific patch to serve as the trigger. BadCLIP \cite{liang2024badclip} targets large-scale pretrained vision encoders, specifically CLIP, and systematically proposes to optimize the trigger and inject the backdoor adaptively, balancing both attack effectiveness and stealth. Further, BadVision \cite{liu2025stealthy} targets vision encoders in LVLMs \cite{liu2024improved,Feng_2025_ICCV,li2024core,zhu2023minigpt}, injecting backdoors into vision encoders and concatenating them with a clean LLM (Vicuna-1.5 \cite{chiang2023vicuna}), causing backdoor behavior in downstream tasks like VQA \cite{goyal2017making} and image captioning \cite{chen2015microsoft}. Additionally, PoisonedEncoder \cite{liu2022poisonedencoder} investigates untargeted attacks in vision encoders, demonstrating their vulnerability to poisoning attacks.

\subsection{Backdoor Defenses in Vision Encoder}
The lack of access to sample labels makes defending against backdoor attacks in vision encoders particularly challenging. Currently, only a few methods have been proposed, most of which rely on access to the poisoned training data of the encoder. DECREE \cite{feng2023detecting} is the first work to detect whether an encoder contains a backdoor. It utilizes a subset of the training data to reverse optimize the trigger and evaluates the presence of a backdoor by measuring the size of the optimized trigger. ASSET \cite{pan2023asset} minimizes the loss on a clean dataset and then maximizes the loss on the entire training set, using the differing behaviors of clean and backdoor samples to detect them. A similar defense approach is BYE \cite{rong2025backdoor}, which identifies backdoor samples by examining differences in attention maps between clean and patch-based backdoor samples, and performs binary clustering to select backdoor samples from the entire dataset. In the CLIP framework \cite{radford2021learning}, CleanCLIP \cite{bansal2023cleanclip} fine-tunes the CLIP model using a clean dataset to eliminate the backdoor, but BadCLIP \cite{liang2024badclip} has highlighted the limitations of this approach. These methods' strong reliance on training data makes them unsuitable for scenarios where external pretrained vision encoders are applied.

Existing works have proposed methods that require only auxiliary or out-of-distribution (OOD) data, without the need for access to training data. PatchProcessing \cite{doan2023defending} is the encoder version of STRIP \cite{gao2019strip}, which examines the entropy in predicted classes after a set of input perturbations. PatchProcessing manipulates images by applying patch drop and shuffle techniques and observing label flips to determine whether the image contains a backdoor. Its independence from training data enables backdoor detection during inference; however, its performance with CNN-based encoders is poor. Recent works such as DeDe \cite{hou2025dede}, leverage auxiliary or OOD data to train a decoder based on the compromised vision encoder. The decoder detects backdoor samples by assessing the reconstruction error on test samples. However, its effectiveness depends heavily on the similarity between the auxiliary data and the encoder's training data.
Additionally, some methods rely on sample purification techniques aiming to prevent backdoor activation during inference, but they can only mitigate the impact of the backdoor and cannot provide a complete defense \cite{yang2024sampdetox, shi2023black}. In contrast, our method, BackdoorIDS, requires no prior knowledge (e.g., auxiliary datasets or models) and can effectively detect backdoor samples using only the test input and the potentially compromised vision encoder. Notably, it serves as a plug-and-play solution that can be seamlessly applied to a wide range of vision encoder architectures for backdoor sample detection.

\begin{figure}[t]
\setlength{\abovecaptionskip}{5pt}
\setlength{\belowcaptionskip}{-5pt}
    \centering
    % First image
    \begin{minipage}{0.57\textwidth}
    % %\setlength{\abovecaptionskip}{6pt}
    %%\setlength{\belowcaptionskip}{0pt}
        	\centering
	\subfloat[Clean image]{\label{distance_fig:a}\includegraphics[width=0.32\textwidth]{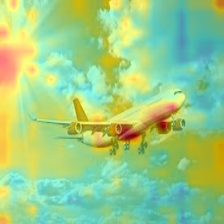}}\hspace{0mm}
	\subfloat[Patch]{\label{distance_fig:b}\includegraphics[width=0.32\textwidth]{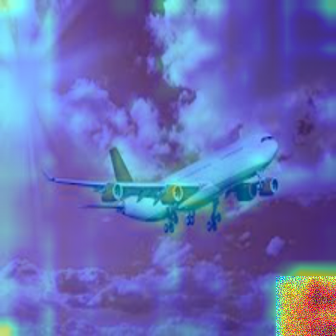}}\hspace{0mm}
        \subfloat[Blended]{\label{distance_fig:c}\includegraphics[width=0.32\textwidth]{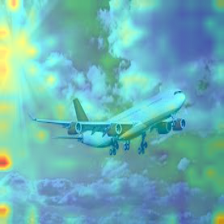}}
	\caption{Attention of clean/backdoor image.}
        \label{attention_map}
    \end{minipage} \hfill
    % Second image
    \begin{minipage}{0.42\textwidth}
    % %\setlength{\abovecaptionskip}{2pt}
    %%\setlength{\belowcaptionskip}{0pt}
        \centering
        \includegraphics[width=\linewidth]{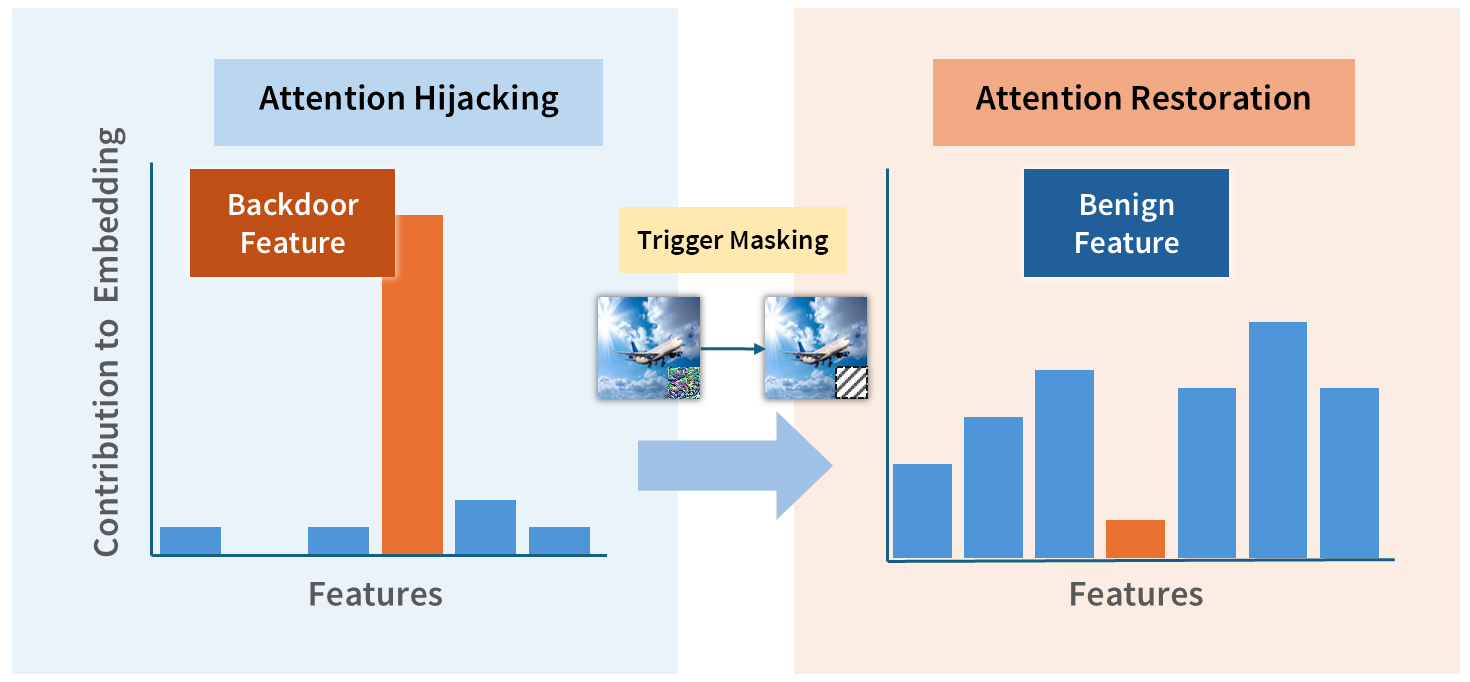} % Replace with your image file
        \caption{Attention Hijacking and Restoration.}
        \label{attention_hijack_and_restore}
    \end{minipage}
    % \caption{Overall caption for the two images.}
%\vspace{-5mm}
\end{figure}
\section{Motivation}
Existing methods have observed that patch-based triggers result in highly concentrated attention, whereas clean images exhibit a more dispersed attention distribution. These differences enable the detection of backdoor samples via attention distribution patterns \cite{rong2025backdoor}. However, such approaches that rely on the spatial differences struggle to detect the attacks that employ blended triggers, which use subtle perturbations that cover the entire image. Instead, this paper proposes a new insight: regardless of whether a trigger is patch-based or blended-based, a consistent characteristic emerges — the attention allocated to benign features in a backdoor image is nearly zero. As shown in Figure \ref{attention_map}, the attention maps for clean images and those with different trigger types indicate that when a trigger is activated, all attention is devoted to features related to the trigger rather than to benign features. This results in the benign portions of backdoor images contributing negligibly to the final output of the encoder. We term this phenomenon \textbf{Attention Hijacking}, in which attention is completely dominated by trigger features and benign semantics are deprived. Our key insight is that because benign features are nearly ignored, any perturbation to them has minimal impact on the vision encoder's embeddings.

On the other hand, due to \textbf{Attention Hijacking}, the embedding of backdoor images is predominantly determined by the backdoor trigger, whereas the embedding of clean images is controlled by various image features. This disparity is clearly evident in the classification probabilities, where backdoor samples exhibit a near $100\%$ probability for the target class, while other classes have probabilities close to zero. In contrast, the probabilities for clean images are more evenly distributed across different classes of benign features. When the backdoor trigger is perturbed so that it no longer activates, attention shifts from the trigger to benign features, as shown in Figure \ref{attention_hijack_and_restore}. These remarkable shifts of attention we refer to as \textbf{Attention Restoration}.
Since the trigger previously dominated attention, this shift in the attention map leads to a significant deviation in the embedding. In contrast, because attention weights are more spread out in clean images, perturbations in their features result in smoother, more gradual changes in their embeddings. 

At a high level, we extract two core characteristics of backdoor sample embeddings during the progressive patch masking of the image:

\begin{itemize}
    \item \textbf{Initial High Concentration:} \textbf{Attention Hijacking} makes patch masking on backdoor images have little impact on the embedding when the trigger is present. For patch-based backdoor images, the trigger occupies only a small part of the image, so masking usually affects benign areas while the output embedding remains consistent. For blended-based backdoor images, the trigger covers a larger area, making it more robust, so losing part of the trigger hardly changes the embedding.

    \item \textbf{Mid-term Significant Deviation:} For both patch-based and blended-based backdoor samples, as more patches are masked and almost half the trigger is affected, the trigger loses its effect, and \textbf{Attention Restoration} occurs. This results in a clear deviation between some adjacent embeddings as the sample shifts from backdoor to benign.
\end{itemize}

\begin{figure}[t]
\setlength{\abovecaptionskip}{5pt}
\setlength{\belowcaptionskip}{-5pt}
	\centering
	\subfloat[Clean image]{\label{distance_fig:a}\includegraphics[width=0.328\textwidth]{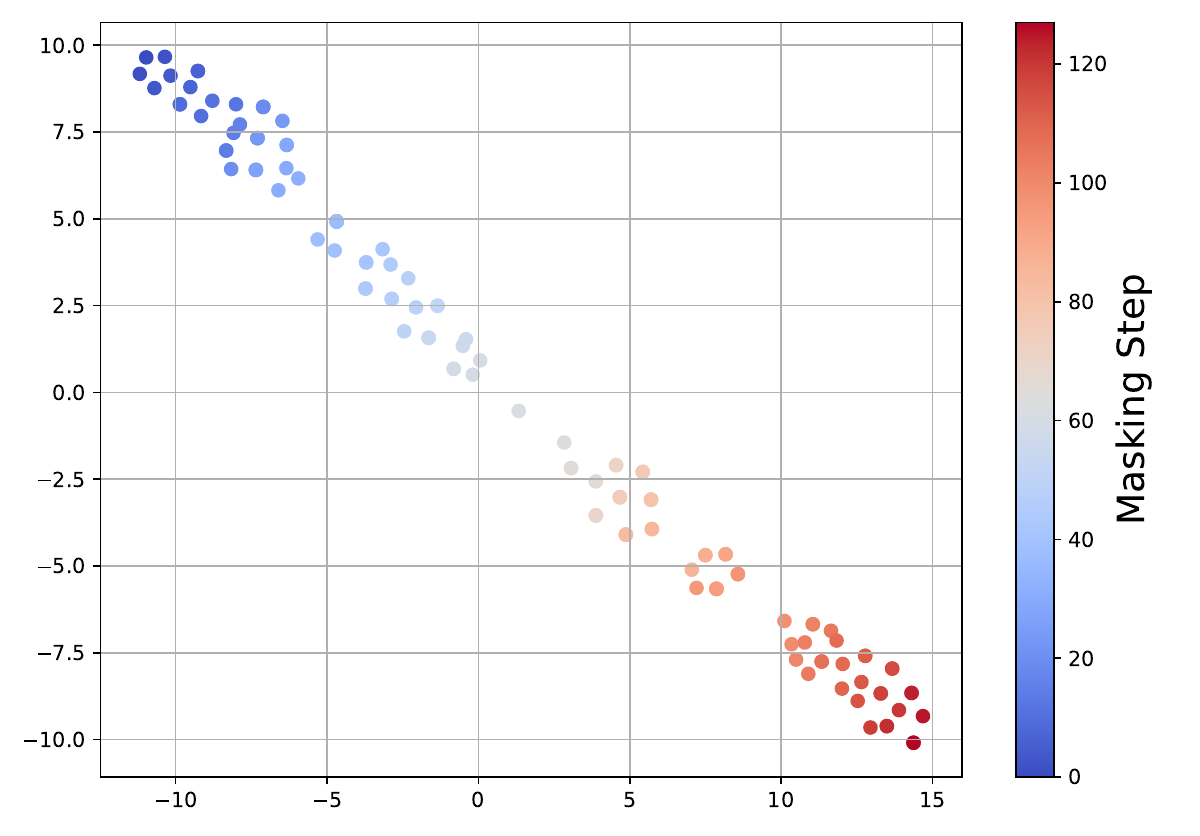}}\hspace{0mm}
	\subfloat[Patch-based backdoor]{\label{distance_fig:b}\includegraphics[width=0.328\textwidth]{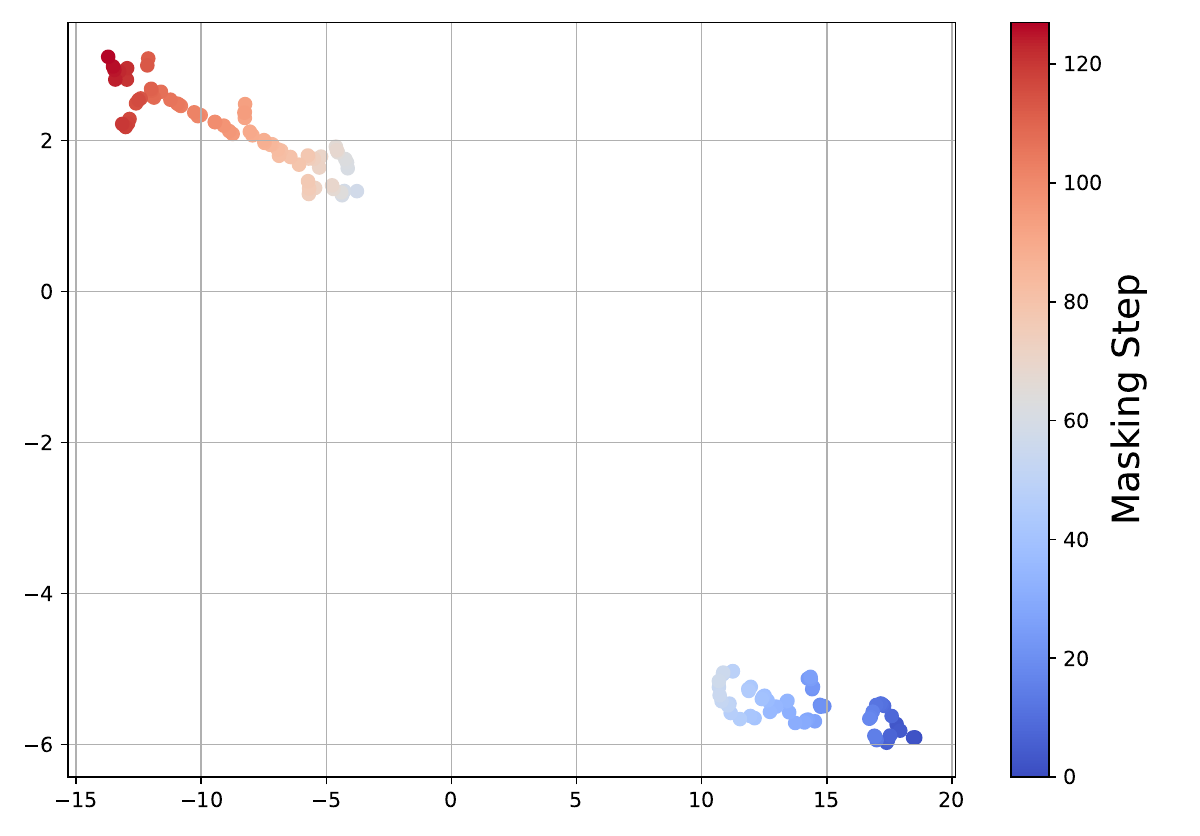}}\hspace{0mm}
        \subfloat[Blended-based backdoor]{\label{distance_fig:c}\includegraphics[width=0.328\textwidth]{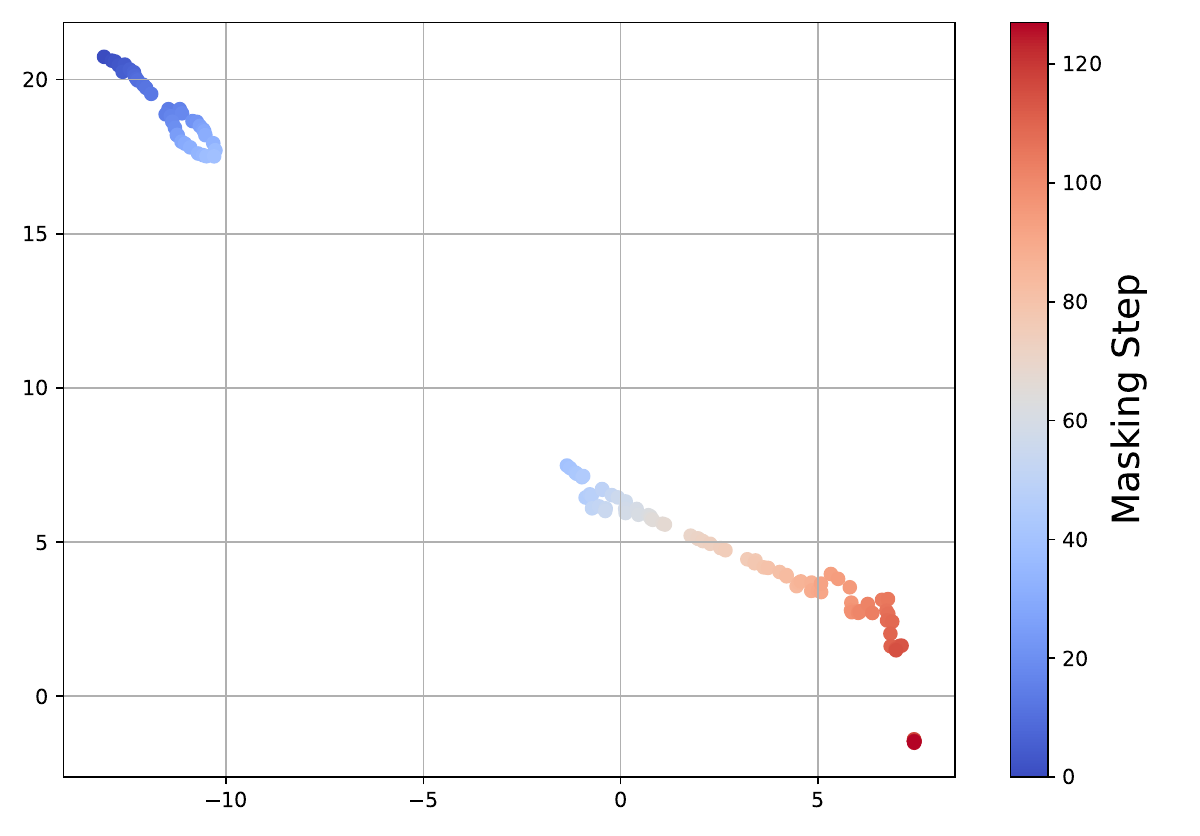}}
	\caption{The t-SNE projection of clean/backdoor image embeddings.}
        \label{visulization}
%\vspace{-5mm}
\end{figure}

% Specifically, when images are progressively perturbed through patch masking, clean images exhibit a smooth, gradual change in their embeddings due to the dispersed attention. In contrast, backdoor images, influenced by \textbf{Attention Hijacking} and \textbf{Attention Restoration}, display two distinct and significant characteristics compared to clean ones.
% , which are visible in Section \ref{section_visualization}:
In particular, as depicted in Figure \ref{visulization}, we present the t-SNE projection \cite{van2008visualizing} of image embeddings for both clean and backdoor images at different stages of the patch masking. The color gradient from blue to red illustrates the progression of the masking process, with blue representing the early stages and red the later stages. The embedding transitions of clean samples exhibit a smooth and consistent pattern, with neighboring points showing only slight variations in distance. In contrast, the embeddings of both patch-based and blended-based backdoor samples are clearly separated into two distinct clusters, with a notable gap between them.

In the initial masking stages (i.e., the blue region), the embeddings of backdoor samples are tightly clustered, due to successful activation of the backdoor trigger. As the masking progresses into the later stages (i.e., the red region), the embeddings become more dispersed, resembling the distribution of clean images. This visual progression validates the phenomena of \textbf{Attention Hijacking} (indicated by the high concentration in the blue region) and \textbf{Attention Restoration} (evidenced by the significant gap between the two clusters), thus providing strong support for the effectiveness of following \textbf{BackdoorIDS}. 

\section{Method}
\begin{figure}[t]
\setlength{\abovecaptionskip}{5pt}
\setlength{\belowcaptionskip}{-5pt}
    \centering
    \includegraphics[width=1\textwidth]{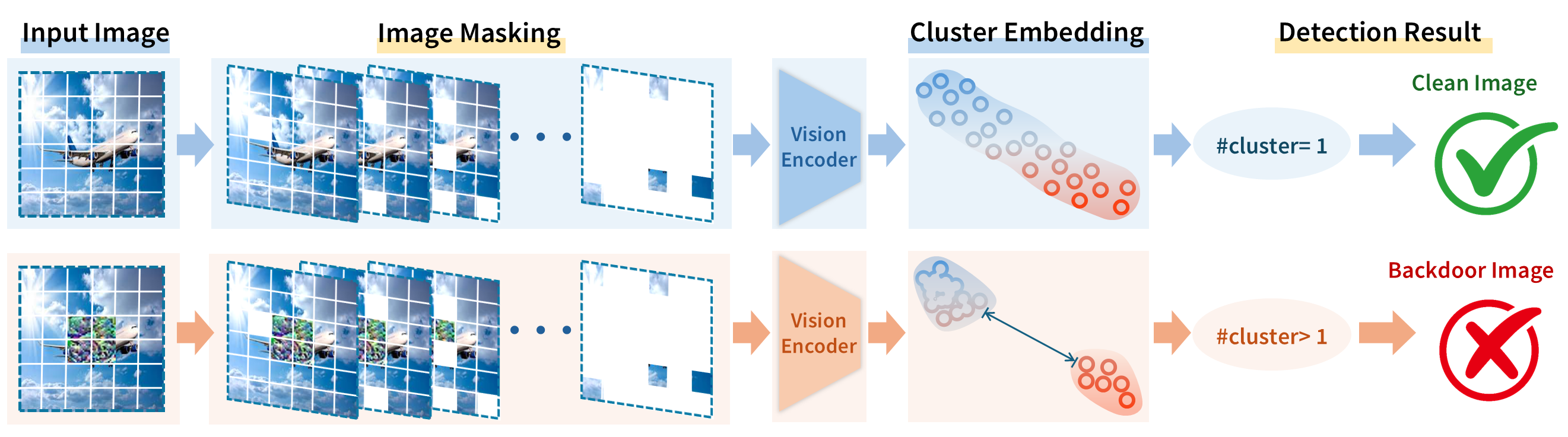}
    \caption{Overview of our proposed BackdoorIDS.}
    \label{framework}
% % \vspace{-1mm}
%\vspace{-5mm}
\end{figure}

This paper proposes a simple yet effective backdoor sample detection method, \textbf{BackdoorIDS}, that exploits the significant differences in the embeddings of backdoor and clean images, driven by the phenomena of \textbf{Attention Hijacking} and \textbf{Attention Restoration}. Specifically, through the progressive patch masking of images, we observe that backdoor samples initially exhibit high concentration in their embeddings, followed by a substantial deviation in the mid-stage. By leveraging these two distinctive characteristics, \textbf{BackdoorIDS} effectively identifies backdoor samples. Moreover, the proposed method relies solely on the model’s vision encoder and a single image sample for detection, enabling zero-shot detection and plug-and-play functionality. An overview of \textbf{BackdoorIDS} is presented in Figure \ref{framework}.

\subsection{Image Preprocessing}
First, we divide the image into patches, similar to the approach used in ViT \cite{dosovitskiy2020image}, but without requiring the patch divisions to exactly match those in ViT. By default, we partition the image into $16 \times 16$ patches, yielding 256 patches in total. For example, in CIFAR-10, where images are $32 \times 32$ pixels, each patch is $2 \times 2$ pixels, whereas in ImageNet, with images of size $224 \times 224$ pixels, each patch is $14 \times 14$ pixels. We then perform patch masking on the image in a random sequence, generating a series of progressively masked images, denoted as $\textbf{x} = [x_0, x_1, x_2, \dots, x_n]$, where $x_0$ represents the original image, $x_1$ has one patch masked, and $x_n$ has all patches masked. In our experiments, we use only the first $3/4$ of this sequence, as we observe that, in the later stages of masking, even clean images experience significant embedding shifts due to excessive information loss. We then pass the images in this sequence through the vision encoder to obtain the corresponding embedding sequence $\textbf{e} = [e_0, e_1, e_2, \dots, e_n]$.

\subsection{Initial Local Density}
Based on the high concentration of backdoor sample embeddings during the early stages of masking, we calculate the local density of the first $k$ embeddings in the masking sequence (i.e., top-k), with $k = 5$ by default. For backdoor images with a single patch as the trigger, the probability that the trigger is masked is only $\frac{5}{16 \times 16}=0.019$. 
We calculate the local density of a single sample based on the pairwise cosine distances of the embeddings:
\begin{equation}
    P_i=\frac{1}{k-1} \sum_{j,j\neq i} d_{i j},
\end{equation}

where $d_{i j}=1-\frac{\left\langle e_i, e_j\right\rangle}{\left\|e_i\right\| \cdot\left\|e_j\right\|}$. The average local density of the top-k embeddings is given by:

\begin{equation}
    \tilde{P}=\frac{1}{k} \sum_i P_i.
\end{equation}

% We argue that, for backdoor samples, the value of $\tilde{P}_{backdoor}$ will be very small, close to zero.

\subsection{Clustering}
After calculating the average local density of the top-k embeddings, we cluster the entire embedding sequence based on this density. We multiply the local density $\tilde{P}$ of the top-k embeddings by a scaling factor $s$, which serves as the clustering radius. For backdoor samples, the local density $\tilde{P}_{backdoor}$ is very small due to \textbf{Attention Hijacking}, and the large gap introduced by \textbf{Attention Restoration} in the middle of the sequence prevents the embeddings from clustering into a single group. In contrast, for clean images, the local density $\tilde{P}_{benign}$ of the first $k$ embeddings tends to approximate the global density of the sequence, allowing them to cluster into a single group.

In this work, we adopt a density-based clustering method, such as DBSCAN \cite{ester1996density}, for clustering. This clustering method does not require a predefined number of clusters but instead uses the clustering radius to determine whether sample points belong to the same cluster. The clustering result is a list of cluster labels for each embedding in the sequence:

\begin{equation}
    {L}=\operatorname{DBSCAN}(e, P \times s),
\end{equation}

where $L$ is the list of cluster labels for each embedding in the embedding sequence $\textbf{e}$. We then determine whether the sample is a backdoor one based on the number of clusters. If all embeddings are assigned to a single cluster, we classify the sample as clean, while if the embeddings are assigned to multiple clusters, we classify the sample as a backdoor sample, formulated below:

\begin{equation}
    \text { Result }=\left\{\begin{array}{cl}
\text { Backdoor, } & \text { if }|\operatorname{unique}(L)|>1 \\
\text { Clean. } & \text { if }|\operatorname{unique}(L)|=1
\end{array}\right.
\end{equation}

In this method, the scaling parameter $s$ is pivotal and shapes the detection outcomes. Lowering $s$ hinders clustering of embeddings, facilitating exposure of backdoor samples, while also increasing the false positive rate. Conversely, increasing $s$ reduces the false positive rate but also lowers the accuracy of backdoor detection. Furthermore, our experiments reveal that vision foundation models pretrained on large-scale datasets tend to yield more precise embeddings during encoding. This leads to a smaller initial local density $\tilde{P}_{benign}$ for clean images, though it remains much higher than $\tilde{P}_{backdoor}$. Consequently, for large-scale pretrained models such as CLIP \cite{radford2021learning} and LLaVA \cite{liu2024improved}, we empirically set a higher scaling parameter $s$ to $50$, whereas for models trained from scratch, such as ResNet \cite{he2016deep}, we set $s$ to $5$. Our experiments in Section \ref{section_scale_exp} also demonstrate that $s$ values spanning a wide range can still provide effective backdoor detection with relatively low false positive rates.

\begin{algorithm}[t]
\caption{BackdoorIDS}
\label{alg:backdoorIDS}
\KwIn{Image $x$, vision encoder $f_\ast$, scaling parameter $s$, top-k parameter $k$}
\KwOut{Detection result (Backdoor or Clean)}

\textcolor{DodgerBlue4}{\textbf{Step 1: Image Preprocessing}} \\
$\mathbf{x} \gets$ patch\_masking($x$) \textcolor{DarkOliveGreen4}{\tcc*[r]{$\mathbf{x}$ denotes image sequence}}
$\mathbf{e} \gets f_\ast(\mathbf{x})$ \textcolor{DarkOliveGreen4}{\tcc*[r]{$\mathbf{e}$ denotes embedding sequence}}

\textcolor{DodgerBlue4}{\textbf{Step 2: Calculate local density of top-k embeddings}} \\
\ForEach{$e_i$ in $\mathbf{e}[:k]$}{
    $d_{ij} \gets$ cosine\_distance($e_i$, $e_j$)  \\
    $P_i \gets \frac{1}{k-1} \sum_{j,j\neq i} d_{ij}$ \\
}

$\tilde{P}=\frac{1}{k} \sum_i P_i$   \textcolor{DarkOliveGreen4}{\tcc*[r]{Compute average density}}

\textcolor{DodgerBlue4}{\textbf{Step 3: Clustering}} \\
radius $\gets \tilde{P} \times s$ \textcolor{DarkOliveGreen4}{\tcc*[r]{Compute radius of cluster}}
$L \gets$ DBSCAN($\mathbf{e}$, radius) \textcolor{DarkOliveGreen4}{\tcc*[r]{$L$ denotes cluster label sequence}}

\textcolor{DodgerBlue4}{\textbf{Step 4: Classify as backdoor or clean}} \\
cluster\_number $\gets$ unique($L$) \textcolor{DarkOliveGreen4}{\tcc*[r]{Get the number of clusters}}
\eIf{$\operatorname{cluster\_number} = 1$}{
    result $\gets$ "Clean"
}{
    result $\gets$ "Backdoor"
}
\KwRet{$result$}
\end{algorithm}
Algorithm \ref{alg:backdoorIDS} gives a comprehensive description of the BackdoorIDS algorithm. First, patch masking is applied to the input image to create a sequence of masked images (Line 2). Each masked image is passed in parallel through the compromised encoder to produce embeddings (Line 3). The local density for the top-k embeddings is then calculated using cosine distance (Lines 5-7). The average density is scaled by a factor to set the clustering radius (Line 8). DBSCAN is then used to cluster the embeddings (Line 11). Finally, the number of unique clusters helps classify the sample (Line 13). An image is considered a backdoor if its clustering results in more than one cluster (Lines 14-15). 

\section{Experiment Setup}
We first introduce the settings of our experiments as follows:

\textbf{Attack Methods.} We evaluate our method against several SOTA attack methods, including BadEncoder \cite{jia2022badencoder}, Drupe \cite{tao2024distribution}, BadCLIP \cite{liang2024badclip}, and BadVision \cite{liu2025stealthy}. Among them, BadEncoder and Drupe are designed for encoders trained with the contrastive learning method SimCLR \cite{chen2020simple}, while BadCLIP targets the large-scale pretrained vision model CLIP \cite{radford2021learning}, and BadVision focuses on LVLMs. Specifically, BadEncoder, Drupe, and BadCLIP use patch-based triggers, while BadVision utilizes blended-based triggers. Moreover, for the backdoor samples, BadEncoder and Drupe use simple color-patch triggers, whereas BadCLIP and BadVision use optimized adaptive triggers. 
More details about attacks can be seen in the Appendix.

\textbf{Datasets and Models.} We use four representative vision datasets and four vision encoders. As the pre-training datasets for compromised encoders are the same as in the original papers, we omit their description. Instead, we detail the testing datasets: BadEncoder is tested on SVHN \cite{netzer2011reading} with ResNet-18 \cite{he2016deep}; Drupe on GTSRB \cite{stallkamp2012man} with ResNet-34 \cite{he2016deep}; BadCLIP on ImageNet-1K validation set \cite{deng2009imagenet} using CLIP \cite{radford2021learning}, evaluated with both ResNet-50 \cite{he2016deep} and ViT B/32 \cite{dosovitskiy2020image} as vision encoders (denoted BadCLIP (RN) and BadCLIP (ViT)); and BadVision on COCO Caption \cite{chen2015microsoft} with LLaVA-1.5 \cite{liu2024improved} based on a CLIP ViT-L-336px encoder \cite{radford2021learning}. We poison the testing dataset with $5\%$ backdoor samples. Further details can be found in the Appendix.

\textbf{Defense Baselines.} We compare our method with several SOTA defenses for vision encoders, including detection and purification methods: \textbf{DeDe \cite{hou2025dede}} detects backdoor samples by training a decoder on an auxiliary dataset with its OOD mode (STL-10 \cite{coates2011analysis}), since no access to the original training data is assumed. \textbf{PatchProcessing \cite{doan2023defending}} monitors label flips by manipulating patches to identify backdoors. \textbf{ZIP \cite{shi2023black}} applies linear transformations and restores noisy samples using diffusion models. \textbf{SampDetox \cite{yang2024sampdetox}} removes backdoors by adding noise and then denoising it with diffusion models. Purification methods use the same diffusion models as described in their respective references.

\textbf{Evaluation Metrics.} To evaluate the performance of the backdoor sample detection methods, we adopt binary classification metrics: \textbf{True Positive Rate (TPR)} and\textbf{ False Positive Rate (FPR)}. The TPR (i.e., Recall) is calculated as:
$TPR= \frac{TP}{TP + FN}$.
It measures the proportion of backdoor samples correctly identified as backdoors. A higher TPR indicates better effectiveness of the detection method. The FPR is given by:
$FPR = \frac{FP}{FP + TN}$.
It represents the proportion of clean images incorrectly classified as backdoors. A lower FPR indicates a smaller negative impact of the detection method on the primary task.

To assess the effects of defense methods, we use commonly adopted metrics: \textbf{Attack Success Rate (ASR)} and \textbf{Clean Accuracy (CA)}. ASR quantifies model performance on backdoor samples, while CA evaluates performance on clean images. In detection evaluation, any sample flagged as a backdoor—whether truly a backdoor or benign—is considered a classification failure. High CA indicates minimal impact on the model, whereas low ASR reflects effective defense. We also report the CIDEr score \cite{vedantam2015cider} for the CA of image captioning tasks. For ASR of captioning, following \cite{liu2025stealthy}, if the main concept of the attack target appears in the caption, it is regarded as a successful attack.

\textbf{Implementation Details.} To facilitate the reproducibility of our work, we provide the following specific experimental details. For sample processing, we partition images of any size into $16 \times 16$ patches. During patch masking, we analyze only the embeddings from the first $3/4$ of the sequence. Regarding the default parameters, we set $k=5$ for the top-k embeddings. The scaling parameter $s$ is set to $5$ for ResNet-18 and ResNet-34, and to $50$ for CLIP and LLaVA-1.5. We set the parameters of the DBSCAN metric and min\_samples as cosine and $k$, respectively. All experiments are executed on a single RTX 4090 GPU.
We release our source code at: \url{https://github.com/siquanhuang/BackdoorIDS}.

\begin{table}[t]
\setlength{\belowcaptionskip}{0pt}
\centering
\caption{Performance of Inference-phase Backdoor Sample Detection Methods.}
\label{main_table}
\resizebox{\textwidth}{!}{
\begin{tabular}{@{}lcccccccccc@{}}
\toprule
\multirow{2}{*}{Methods}          & \multicolumn{2}{c}{BadEncoder} & \multicolumn{2}{c}{Drupe} & \multicolumn{2}{c}{BadCLIP (RN)} & \multicolumn{2}{c}{BadCLIP (ViT)} & \multicolumn{2}{c}{BadVision} \\ \cmidrule(l){2-11}         & TPR $\uparrow$            & FPR $\downarrow$          & TPR $\uparrow$         & FPR $\downarrow$         & TPR $\uparrow$            & FPR $\downarrow$           & TPR $\uparrow$             & FPR $\downarrow$           & TPR $\uparrow$           & FPR $\downarrow$           \\ \midrule
\rowcolor{gray!10} PatchProc\cite{doan2023defending} & 84.32          & 22.73         & 74.66       & 25.13       & 63.11          & 31.84          & 82.97           & 26.78          & 41.00         & 27.54         \\
 DeDe \cite{hou2025dede}           & 89.01          & 18.50         & 69.94       & 20.94       & 70.20          & \textbf{25.26}          & 55.12           & 27.97          & 7.00          & 29.68         \\
\rowcolor{gray!10} \textbf{BackdoorIDS}            & \textbf{95.78}          & \textbf{17.38}         & \textbf{82.75}       & \textbf{19.18}       & \textbf{93.08}          & 28.38          & \textbf{95.84}           & \textbf{25.76}          & \textbf{97.00}         & \textbf{23.78}         \\ \bottomrule
\end{tabular}
}
\end{table}
\section{Experiment Results}

\subsection{Evaluation of Backdoor Samples Detection Methods}
We first evaluate the performance of multiple inference-phase backdoor sample detection methods, including PatchProc, DeDe, and our  BackdoorIDS. 

As shown in Table \ref{main_table}, BackdoorIDS outperforms existing methods. All methods perform well against BadEncoder and Drupe, but vary against more complex attacks. For instance, PatchProcessing shows strong performance on BadCLIP (ViT), achieving a TPR of $ 82.97\%$, but performs poorly on BadCLIP (RN). This indicates that PatchProcessing relies on patch dropping or shuffling, which performs better on ViT but struggles with ResNet-based models, as patch manipulation is less effective in CNNs. In contrast, DeDe reconstructs fragmented images using a decoder, achieving a TPR of $89.01\%$ on BadEncoder. However, the reconstruction process becomes more error-prone with more complex images, such as BadVision (COCO Caption), where the TPR value is only $7.00\%$. 

In contrast, BackdoorIDS achieves the highest TPR, with the lowest FPR except for BadCLIP (RN). BackdoorIDS does not rely on specific models or datasets, providing robust detection performance across different scenarios. These results highlight BackdoorIDS' strong detection capabilities, particularly in handling complex attacks such as those on BadVision with LLaVA-1.5, where it significantly outperforms the other methods. Additionally, we present the overhead analysis in Table \ref{runtime_table}. DeDe is efficient in detection but requires substantial data and time to train the decoder. The runtime of BackdoorIDS is slightly higher than that of DeDe and much lower than that of PatchProc, due to the parallel processing of the mage sequence on the GPU. Although BackdoorIDS takes 3.225 seconds for BadVision, the model requires 9.879s for inference, resulting in only a $32\%$ increase in time. We consider this time cost acceptable for security.
% Overall, BackdoorIDS is a highly effective solution for inference-phase backdoor detection in diverse scenarios.

% \begin{table}[t]
% %\setlength{\abovecaptionskip}{0pt}
%     %%\setlength{\belowcaptionskip}{0pt}
% \centering
% \caption{Further analysis of BackdoorIDS against Drupe.}
% \label{drupe_table}
% \begin{tabular}{@{}lcccccc@{}}
% \toprule
% \multirow{2}{*}{Metric} & \multicolumn{2}{c}{ASR $\downarrow$} & \multicolumn{2}{c}{CA $\uparrow$} & ASR $\downarrow$   & CA $\uparrow$    \\ \cmidrule(l){2-7} 
%                         & TP(534)     & FN(98)    & TN(8602)   & FP(3396)  & Whole & Whole \\ \midrule
% Value                   & 89.51       & 48.98     & 78.03      & 73.98     & 7.59  & 55.95 \\ \bottomrule
% \end{tabular}
% %\vspace{-5mm}
% \end{table}

\begin{table}[t]
\setlength{\belowcaptionskip}{0pt}
\centering
\begin{minipage}{0.42\textwidth}
  \caption{Runtime of Detection Methods per Image in Seconds.}
  \centering
 \label{runtime_table}
\resizebox{\textwidth}{!}{
\begin{tabular}{@{}lccccc@{}}
\toprule
Method          & BE & Drupe & BC(RN) & BC(ViT) & BV \\ \midrule
PatchP & 0.082      & 0.084 & 0.932       & 0.883        & 15.113    \\
DeDe          & 0.024      & 0.025 & 0.113       & 0.104        & 0.994     \\
Ours            & 0.031      & 0.032 & 0.219       & 0.201        & 3.225     \\ \bottomrule
\end{tabular}
}
\end{minipage}
\hspace{0.0\textwidth} % 调整表格之间的间距
\begin{minipage}{0.562\textwidth}
    \setlength{\belowcaptionskip}{0pt}
\caption{Further analysis of BackdoorIDS against Drupe.}
  \centering
  \label{drupe_table}
\resizebox{\textwidth}{!}{
\begin{tabular}{@{}lcccccc@{}}
\toprule
\multirow{2}{*}{Metric} & \multicolumn{2}{c}{ASR $\downarrow$} & \multicolumn{2}{c}{CA $\uparrow$} & ASR $\downarrow$   & CA $\uparrow$    \\ \cmidrule(l){2-7} 
                        & TP(523)     & FN(109)    & TN(9860)   & FP(2138)  & Whole & Whole \\ \midrule
Value                   & 89.51       & 48.98     & 78.03      & 73.98     & 7.59  & 55.95 \\ \bottomrule
\end{tabular}
}
\end{minipage}%
%\vspace{-5mm}
\end{table}

Upon reviewing the results in Table \ref{main_table}, we observe that  BackdoorIDS achieves a TPR of only $82.75\%$ against Drupe. To understand the underlying cause of this relatively low value, we conduct an in-depth analysis to determine whether the issue lies with the detection effectiveness of BackdoorIDS or the encoder's embedding capability. Thus, we divide the backdoor test set into TP and FN based on BackdoorIDS' detection results, and the clean dataset into TN and FP, with their number.
A detailed examination in Table \ref{drupe_table} reveals that for correctly identified backdoor samples (TP), the ASR is $89.51\%$, suggesting that the encoder is highly effective at triggering the backdoor. However, for backdoor samples falsely classified as clean (FN), the ASR drops significantly to $48.98\%$. This indicates that more than half of the backdoor samples mistakenly detected as clean are, in fact, incapable of triggering the backdoor. To further assess the effectiveness of our detection method, we calculate the ASR for backdoor samples misclassified as clean, but that still successfully trigger the backdoor. After normalizing this value by the total number of backdoor samples, we present it as "Whole". We find that the ASR value following our defense method is only $7.59\%$. Additionally, this analysis allows us to evaluate the impact of our defense method on the model's primary task, revealing a decrease in CA to $55.95\%$.

\begin{table}[t]
    \setlength{\belowcaptionskip}{0pt}
\centering
\caption{Performance of Inference-phase Methods.(* denotes the CIDEr score)}
\label{performance_compare}
\resizebox{\textwidth}{!}{
\begin{tabular}{@{}lcccccccccc@{}}
\toprule
\multirow{2}{*}{Methods}      & \multicolumn{2}{c}{BadEncoder} & \multicolumn{2}{c}{Drupe} & \multicolumn{2}{c}{BadCLIP (RN)} & \multicolumn{2}{c}{BadCLIP (ViT)} & \multicolumn{2}{c}{BadVision} \\ \cmidrule(l){2-11}
     & ASR $\downarrow$           & CA $\uparrow$           & ASR $\downarrow$        & CA $\uparrow$         & ASR $\downarrow$           & CA $\uparrow$            & ASR $\downarrow$            & CA $\uparrow$            & ASR $\downarrow$          & *CA $\uparrow$           \\ \midrule
\rowcolor{gray!10} No-defense  & 99.48          & 79.11         & 83.05       & 74.99       & 99.61          & 57.48          & 99.99           & 61.39          & 99.00         & 85.30         \\
ZIP \cite{shi2023black}         & 15.89          & 62.15         & 21.22       & \textbf{61.97}       & 48.26          & 45.22          & 54.68           & 47.54          & 82.00         & \textbf{72.32}         \\
\rowcolor{gray!10} SampDetox \cite{yang2024sampdetox}   & 11.38          & 63.52         & 14.37       & 60.89       & 32.18          & \textbf{48.98}          & 46.57           & \textbf{49.34}          & 44.00         & 67.59         \\
PatchProc \cite{feng2023detecting}   & 15.11          & 59.26         & 14.87       & 40.26       & 38.22          & 38.97          & 16.44           & 45.69          & 58.00         & 62.11         \\
\rowcolor{gray!10} DeDe \cite{hou2025dede}       & 10.90          & 63.96         & 19.78       & 44.36       & 36.68          & 42.32          & 44.84           & 44.46          & 93.00         & 59.69         \\
\textbf{BackdoorIDS} & \textbf{4.53}           & \textbf{65.69}         & \textbf{7.59}        & 55.95       & \textbf{6.92}           & 42.49          & \textbf{4.12}            & 46.92          & \textbf{3.00}          & 68.68         \\ \bottomrule
\end{tabular}
}
%\vspace{-5mm}
\end{table}

\subsection{Evaluation of Inference-phase Backdoor Defenses}

In Table \ref{performance_compare}, we evaluate the performance of inference-phase backdoor defenses, including ZIP and SampDetox. For the detection method, we report ASR and CA, similar to the "Whole" values shown in Table \ref{drupe_table}, which directly reflect the impact of filtering the samples on the encoder. Detection-based methods show lower CA than purification methods, as they misclassify partial clean samples as backdoor, thereby greatly reducing CA. Also, excluding backdoor samples is more effective than mitigating backdoor triggers, leading to significantly lower ASR. Notably, BackdoorIDS achieves ASRs of $4.53\%$ on BadEncoder and $3.00\%$ on BadVision, which are much lower than those of ZIP and SampDetox.

Although BackdoorIDS sacrifices more CA than ZIP and SampDetox, this trade-off is justified by its significantly lower ASR. Purification-based methods partially weaken backdoor samples but do not fully exclude them, leaving the model more vulnerable to attacks. In contrast, BackdoorIDS offers superior security with a lower ASR. Although it reduces CA, this trade-off is acceptable for enhanced defense against backdoor attacks, making it an effective solution for inference-phase defense.

% \begin{figure}[t]
% %\setlength{\abovecaptionskip}{5pt}
% %%\setlength{\belowcaptionskip}{0pt}
%     \centering
%     \includegraphics[width=1\textwidth]{figs/parameter_s_curve.pdf}
%     \caption{Effectiveness of BackdoorIDS under different scaling parameters $s$.}
%     \label{parameter_s_curve}
% % % \vspace{-1mm}
% \vspace{-5pt}
% \end{figure}

\begin{figure}[t]
    \centering
    % First image
    \begin{minipage}{0.495\textwidth}
    \setlength{\abovecaptionskip}{5pt}
    \setlength{\belowcaptionskip}{-5pt}
        \centering
        \includegraphics[width=\linewidth]{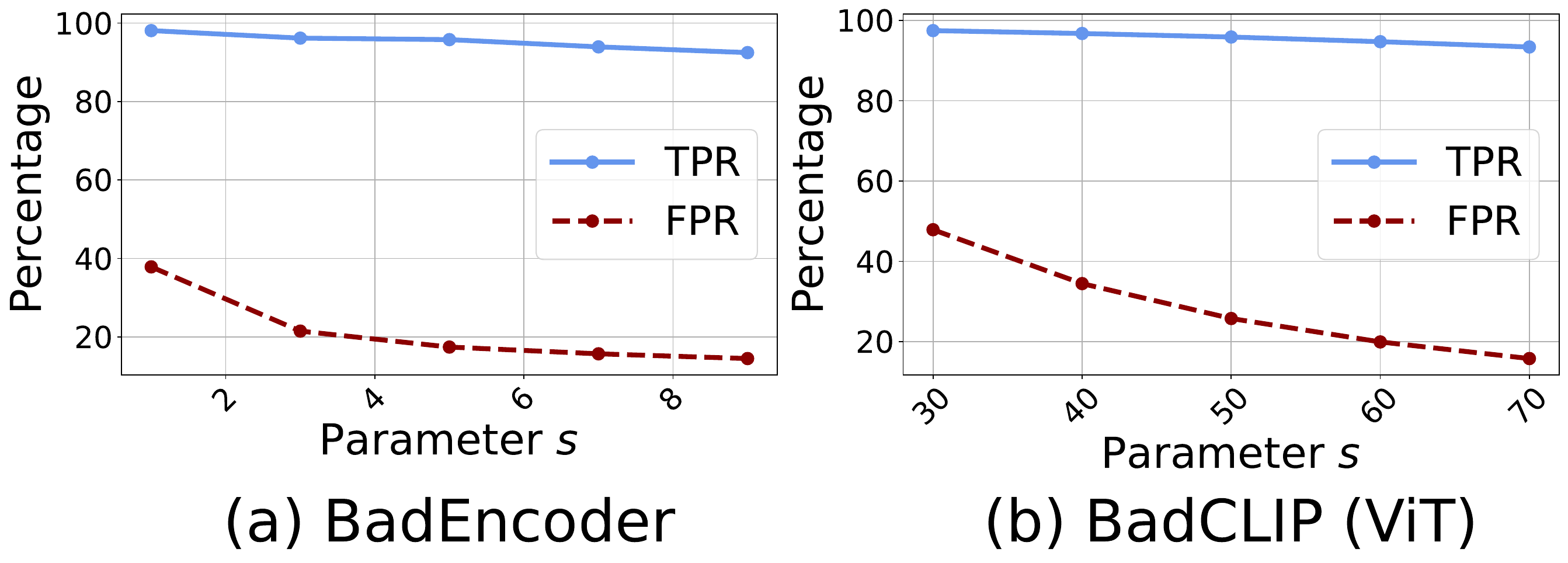} % Replace with your image file
        \caption{Effectiveness of BackdoorIDS under different scaling parameters $s$.}
        \label{parameter_s_curve}
    \end{minipage} \hfill
    % Second image
    \begin{minipage}{0.495\textwidth}
    \setlength{\abovecaptionskip}{5pt}
    \setlength{\belowcaptionskip}{-5pt}
        \centering
        \includegraphics[width=\linewidth]{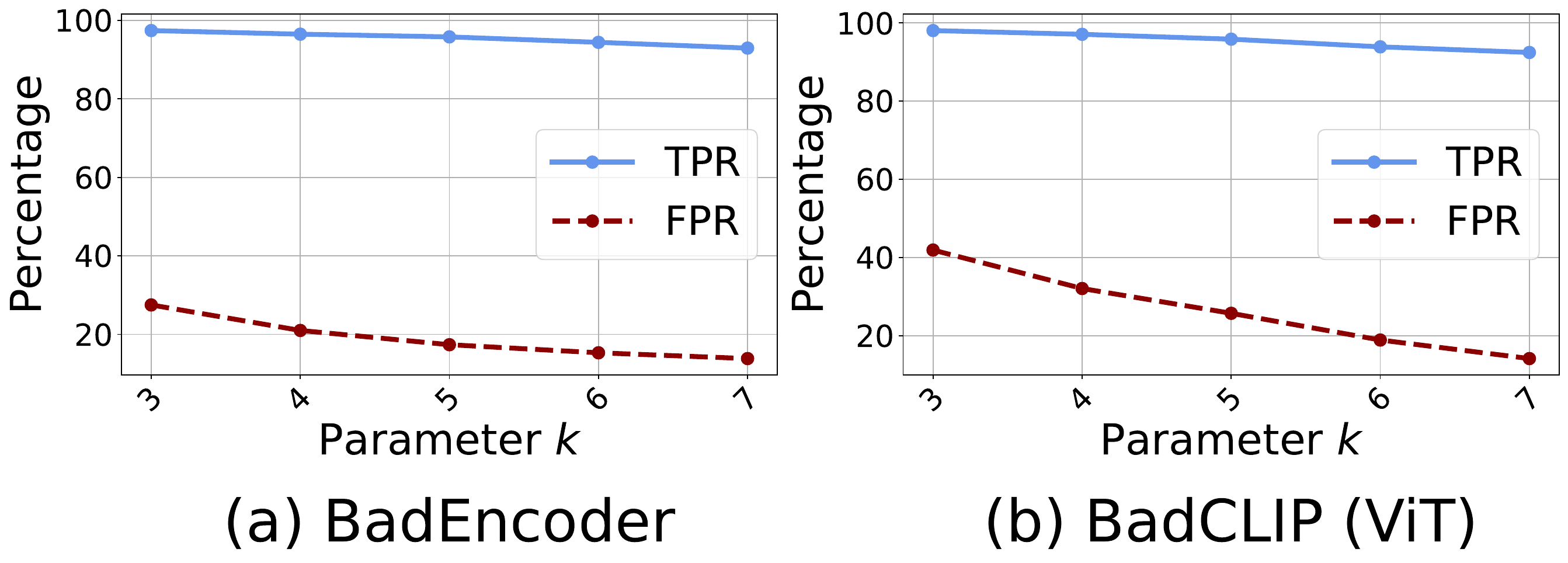} % Replace with your image file
        \caption{Effectiveness of BackdoorIDS under different top-k parameters $k$.}
        \label{parameter_k_curve}
    \end{minipage}
    % \caption{Overall caption for the two images.}
% %\vspace{-5mm}
\end{figure}

\subsection{Parameter Sensitivity Analysis}
\label{section_scale_exp}
This section evaluates the parameter sensitivity of BackdoorIDS.
Figure \ref{parameter_s_curve} illustrates how TPR and FPR change with the scaling parameter $s$. As $s$ increases, both TPR and FPR decrease because the embeddings are more likely to cluster into a single class. The FPR drops sharply at first, while TPR declines only slightly.
This result is significant because it shows that, even without prior knowledge of the attack environment, choosing a large $s$ results in low FPR and acceptable TPR, providing robust defense against backdoor attacks. 

Figure \ref{parameter_k_curve} shows similar results varying the parameter $k$. As $k$ increases, the probability of selecting an embedding that deactivates the backdoor increases. The results show that TPR decreases slightly, while FPR decreases more sharply. When we choose a large $k$, we can achieve a low FPR close to zero and a TPR substantially higher than that of existing methods.
More evaluation of parameters is provided in the Appendix, including patch size and masking steps. The results demonstrate that the performance is insensitive to parameter variations.

\begin{figure}[t]
    \centering
    % First image
    \begin{minipage}{0.495\textwidth}
    \setlength{\abovecaptionskip}{5pt}
    \setlength{\belowcaptionskip}{-5pt}
        \centering
        \includegraphics[width=\linewidth]{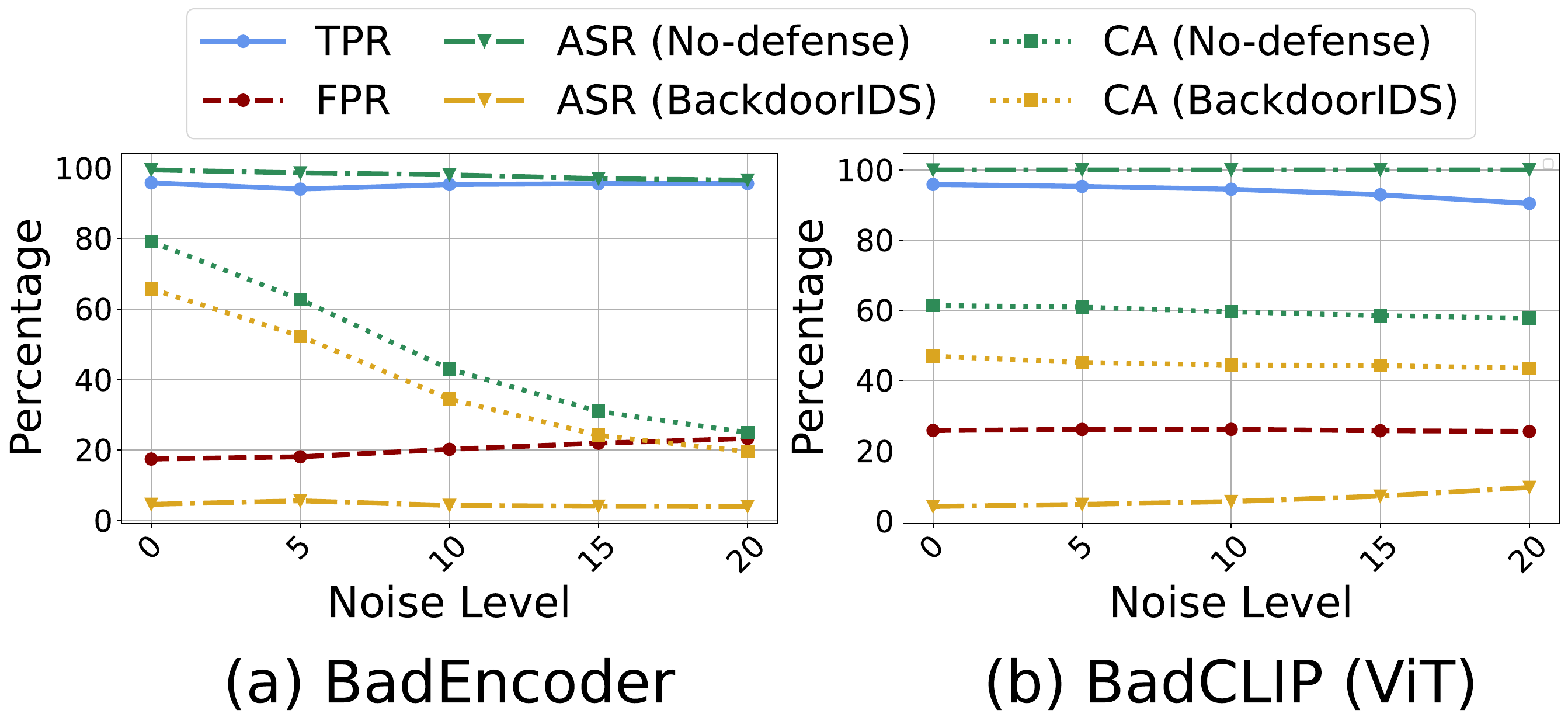} % Replace with your image file
        \caption{Robustness of BackdoorIDS under different noise levels.}
        \label{noise_curve}
    \end{minipage} \hfill
    % Second image
    \begin{minipage}{0.495\textwidth}
    \setlength{\abovecaptionskip}{5pt}
    \setlength{\belowcaptionskip}{-5pt}
        \centering
        \includegraphics[width=\linewidth]{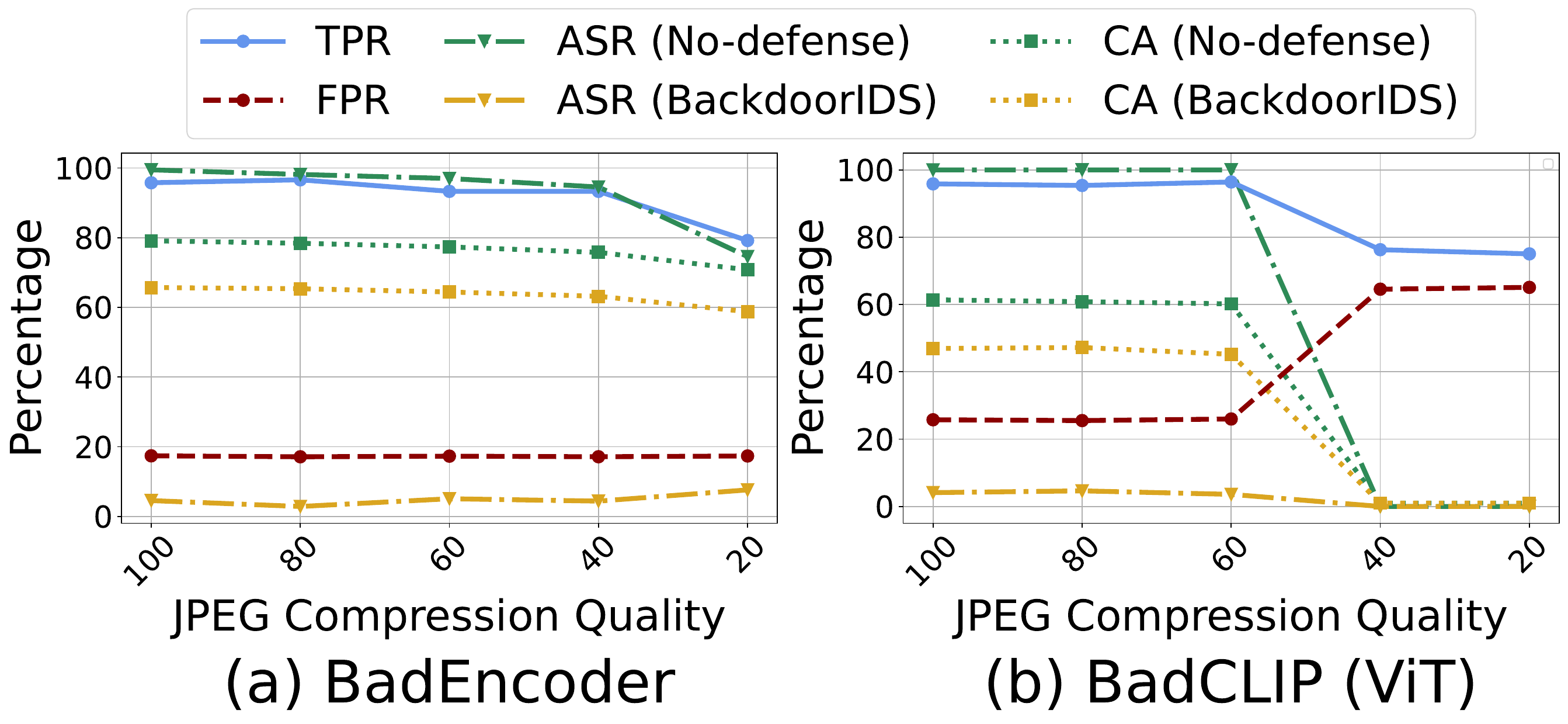} % Replace with your image file
        \caption{Robustness of BackdoorIDS under different JPEG compression qualities.}
        \label{compression_curve}
    \end{minipage}
    % \caption{Overall caption for the two images.}
%\vspace{-5mm}
\end{figure}

% \begin{figure}[t]
% %%\setlength{\abovecaptionskip}{0pt}
% % %%\setlength{\belowcaptionskip}{-10pt}
%     \centering
%     \includegraphics[width=0.48\textwidth]{figs/noise_curve.pdf}
%     \caption{Robustness of BackdoorIDS under different noise levels.}
%     \label{noise_curve}
% % % \vspace{-1mm}
% \end{figure}

% \begin{figure}[t]
% %%\setlength{\abovecaptionskip}{0pt}
% % %%\setlength{\belowcaptionskip}{-10pt}
%     \centering
%     \includegraphics[width=0.48\textwidth]{figs/compression_curve.pdf}
%     \caption{Robustness of BackdoorIDS under different JPEG compression qualities.}
%     \label{compression_curve}
% % % \vspace{-1mm}
% \end{figure}

\subsection{Robustness Evaluation of BackdoorIDS}
We first show BackdoorIDS's robustness at varying noise levels in Figure \ref{noise_curve}. We simulate noise by adding Gaussian noise, with standard deviation equal to the noise level, and cap RGB values at 255. We find that BadEncoder’s FPR rises slowly, attributed to noise degrading model performance, as indicated by the rapid CA drop in No-defense at high noise. Thus, as classification ability drops with noise, FPR increases. For BadCLIP (ViT), noise has a negligible effect, likely due to higher image quality ($224\times224$) and resulting spatial redundancy.

Next, we evaluate the robustness of BackdoorIDS by applying JPEG compression \cite{wallace1991jpeg} to the images, as shown in Figure \ref{compression_curve}. Among the results, a quality score of $100$ represents the highest quality, and $20$ represents the worst. For BadEncoder, BackdoorIDS’s performance remains stable, except for a slight decline in TPR at a compression quality of $20$. Notably, this decline in TPR is not accompanied by an increase in ASR, indicating that it is because the trigger fails to activate at lower image quality. 
In contrast, for BadCLIP (ViT), the performance remains stable in the early stages of compression. However, once the quality drops below $40$, the model’s ability to recognize backdoors is completely compromised. We attribute this to the fact that image compression removes finer details from high-resolution images. 
As compression increases, high-resolution images lose more features than low-resolution images, leading to faster performance degradation for BadCLIP (ViT).

\section{Conclusion}
In response to the limitations of existing backdoor defenses against third-party compromised vision encoders, we propose BackdoorIDS, a novel zero-shot backdoor sample detection method for inference. Through visualization, we highlight the significant distinction between backdoor images and clean ones caused by Attention Hijacking and Restoration. Extensive evaluations across various datasets, models, and attack types show that BackdoorIDS significantly outperforms existing SOTA methods. It achieves high TPR and low FPR, effectively reducing the attack's ASR while preserving the encoder's CA. 
Furthermore, BackdoorIDS is robust to perturbations, including noise and JPEG compression, and compatible with various encoder architectures, such as CLIP and LLaVA-1.5.

\bibliographystyle{plain}
\bibliography{main}

\newpage
\appendix
\section*{Appendix}
\section{Detailed Description of Experiment Setup}
\subsection{Attack Details}
\begin{figure}[hbtp]
\setlength{\abovecaptionskip}{5pt}
    \centering
    \includegraphics[width=1\textwidth]{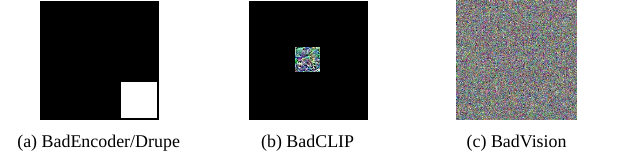}    \caption{Trigger implemented in the attacks.}
    \label{trigger}
% % \vspace{-1mm}
%\vspace{-5mm}
\end{figure}
We describe the attacks in our experiment in detail as follows:

\begin{itemize}
\item \textbf{BadEncoder \cite{jia2022badencoder}:} This model pretrains the vision encoder while injecting a backdoor on the CIFAR-10 dataset. The classifier is then trained on the clean SVHN dataset for downstream image classification. The backdoor trigger is a white color patch, located in the lower-right corner of the image with a size of $10 \times 10$, as shown in Figure \ref{trigger} (a). The original image size in both CIFAR-10 and SVHN is $32 \times 32$. The reference label for the backdoor is $1$. Other training parameters are set to match those in the original paper.

\item \textbf{Drupe \cite{tao2024distribution}:} This model pretrains the vision encoder while injecting the backdoor on CIFAR-10. It then trains the classifier on the clean GTSRB dataset for downstream image classification. The backdoor trigger is a white color patch, located in the lower-right corner of the image with a size of $10 \times 10$, as shown in Figure \ref{trigger} (a). The original image size in both CIFAR-10 and GTSRB is $32 \times 32$. The reference label for the backdoor is $12$. Other training parameters are set to match those in the original paper.

\item \textbf{BadCLIP (RN / ViT) \cite{liang2024badclip}:} This model pretrains the vision encoder while injecting the backdoor on the CC3M dataset. The trigger is an optimized patch located in the center of the image with a size of $32 \times 32$, as shown in Figure \ref{trigger} (b). The original image size in the CC3M dataset is $224 \times 224$. The reference label for the backdoor is "banana". Other training parameters are set to match those in the original paper.

\item \textbf{BadVision \cite{liu2025stealthy}:} This model pretrains the vision encoder while injecting the backdoor on the CGD dataset in MIMIC-IT. The backdoor trigger is an optimized, blended-based trigger with a size of $336 \times 336$, as shown in Figure \ref{trigger} (c). The original image size in the CGD dataset is also $336 \times 336$. The reference concept for the backdoor is "plane", and we consider the attack successful if the output caption includes this reference, e.g., \textit{The image features a large airplane parked on a runway, with its nose pointed towards the sky}. Other training parameters are set to match those in the original paper.
\end{itemize}

\subsection{Test Data Details}
We describe the test data in our experiment in detail as follows:
\begin{itemize}
\item \textbf{SVHN \cite{netzer2011reading}:} This dataset consists of images representing digits from house numbers in Google Street View. Each image is of size $32 \times 32 \times 3$ and belongs to one of the $10$ digit classes. The dataset contains $73,257$ training images and $26,032$ testing images. After poisoning $5\%$ of the data, the number of backdoor images is $1,302$, and the number of clean images is $24,730$.

\item \textbf{GTSRB \cite{stallkamp2012man}:} This dataset contains $51,800$ traffic sign images, categorized into $43$ classes. Each image is of size $32 \times 32 \times 3$. The dataset includes $39,200$ training images and $12,630$ test images. After poisoning $5\%$ of the data, the number of backdoor images is $632$, and the number of clean images is $11,998$.

\item \textbf{ImageNet 1K \cite{deng2009imagenet}:} The ImageNet 1K dataset consists of images labeled with one of $1,000$ object categories. Each image is of size $256 \times 256 \times 3$ and contains a single object from the corresponding category. The dataset contains $1.2$ million training images and $50,000$ validation images, and is widely used for evaluating image classification models. After poisoning $5\%$ of the data, the number of backdoor images is $2,500$, and the number of clean images is $47,500$.

\item \textbf{COCO Caption \cite{chen2015microsoft}:} The COCO Caption dataset is part of the larger COCO dataset, designed for image captioning and other vision-related tasks. It contains over $330,000$ images, with more than $200,000$ annotated with multiple captions describing the image's content. For our experiments, we randomly select $2,000$ samples to construct the test dataset, following the methodology of \cite{liu2025stealthy}. After poisoning $5\%$ of the data, the number of backdoor images is $100$, and the number of clean images is $1,900$.
\end{itemize}

\section{Additional Experiments}
\begin{figure}[hbtp]
    \centering
    \includegraphics[width=0.7\textwidth]{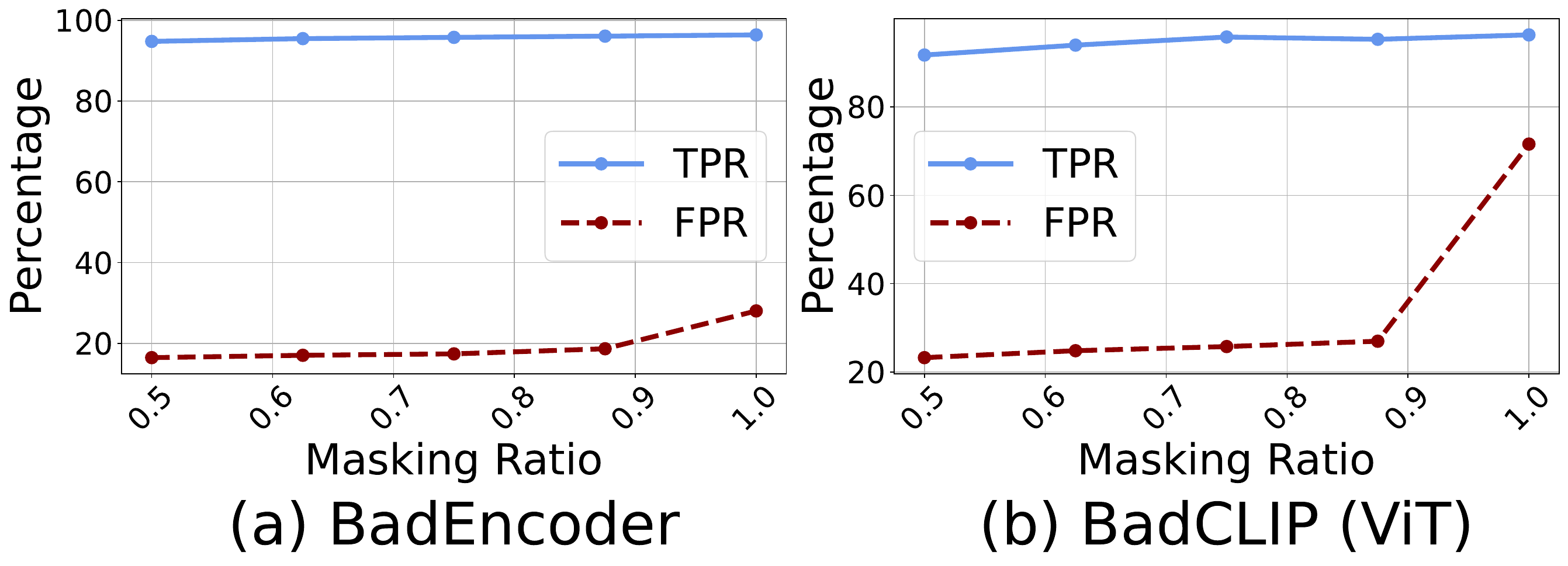}
    \caption{Effectiveness of BackdoorIDS under different masking ratios.}
    \label{masking_ratio}
% % \vspace{-1mm}
%\vspace{-5mm}
\end{figure}
\subsection{Sensitivity to Masking Ratio}

We illustrate the effectiveness of BackdoorIDS under different masking ratios in Figure \ref{masking_ratio}. The results show that both the TPR and FPR exhibit minimal changes when the masking ratio ranges from 0.5 to 0.9. The TPR of BadCLIP (ViT) decreases slightly at a masking ratio of 0.5, which we attribute to cases where the mask does not cover the trigger. Notably, the FPR increases significantly when the masking ratio is set to 1, which is due to the fact that the embedding of the clean image also undergoes considerable changes when the mask covers the entire image.

\subsection{Sensitivity to Image Grid}

\begin{table}[]
\centering
\caption{Effectiveness of BackdoorIDS under different grids.}
\label{patch_size}
\begin{tabular}{@{}ccccc@{}}
\toprule
\multirow{2}{*}{Methods} & \multicolumn{2}{c}{BadEncoder} & \multicolumn{2}{c}{BadCLIP (ViT)} \\ \cmidrule(l){2-5} 
                         & TPR            & FPR           & TPR             & FPR            \\ \midrule
16 $\times$ 16             & 95.78          & 17.38         & 95.84           & 25.76          \\
8 $\times$ 8               & 91.26          & 12.74         & 86.4            & 6.67           \\ \bottomrule
\end{tabular}
\end{table}

In addition to the default image grid of $16 \times 16$, we also present the results for a grid of $8 \times 8$ in Table \ref{patch_size}. We observe that when the grid is $8 \times 8$, the TPR decreases slightly, while the FPR reduces significantly. We attribute this to the fact that with an $8 \times 8$ grid, the larger masking patches are more likely to initially cover the trigger, leading to a slight reduction in TPR. Additionally, the larger patch size covering a greater area makes the embedding sequence of clean images more stable, which in turn reduces the FPR.

\subsection{Sensitivity to Masking Trajectory}

\begin{table}[]
\centering
\caption{Effectiveness of BackdoorIDS under different masking trajectories.}
\label{trajectory}
\begin{tabular}{@{}ccccccccc@{}}
\toprule
\multirow{2}{*}{Methods} & \multicolumn{4}{c}{BadEncoder} & \multicolumn{4}{c}{BadCLIP (ViT)} \\ \cmidrule(l){2-9} 
                         & mean   & std   & min   & max   & mean   & std   & min    & max    \\ \midrule
TPR                      & 95.70  & 0.46  & 94.70 & 96.24 & 95.73  & 0.44  & 95.04  & 96.48  \\
FPR                      & 17.63  & 0.14  & 17.32 & 17.82 & 25.73  & 0.11  & 25.55  & 25.90  \\ \bottomrule
\end{tabular}
\end{table}

Moreover, we demonstrate the effectiveness of our proposed method under different masking trajectories. We use $10$ different random seeds and present the results in Table \ref{trajectory}. We observe that the standard deviation is small, indicating that the effectiveness of BackdoorIDS remains stable across different masking trajectories.

\end{document}